\documentclass[letterpaper, 10 pt, conference]{ieeeconf}

\IEEEoverridecommandlockouts   

\usepackage[utf8]{inputenc} % allow utf-8 input
\usepackage[T1]{fontenc}    % use 8-bit T1 fonts
\usepackage{url}            % simple URL typesetting
\usepackage{booktabs}       % professional-quality tables
\usepackage{amsfonts}       % blackboard math symbols
\usepackage{graphicx}%
\usepackage{amsmath}
\usepackage{amssymb}
\usepackage{xcolor}
\usepackage{accents}
\usepackage{caption}
\usepackage{cite}
\usepackage{cite}
\usepackage{xcolor}
\usepackage{multirow}
\usepackage{siunitx}
\usepackage{amsmath,bm}
\usepackage{algorithm}
\usepackage{algpseudocode}
\usepackage{caption}
\usepackage{subcaption}
\usepackage{stfloats}

\newtheorem{remark}{\textbf{Remark}}

\title{\LARGE \bf
Distributed Invariant Kalman Filter for Cooperative Localization using Matrix Lie Groups
}
\author{Yizhi Zhou, Yufan Liu, Pengxiang Zhu and Xuan Wang
\thanks{Y. Zhou and X. Wang are with the Electrical and Computer Engineering department at George Mason University. Y. Liu is with the Electrical Engineering and Computer Science, University of California, Berkeley, Berkeley, CA, USA. P. Zhu is with the Electrical Engineering and Computer Engineering department at , University of California, Riverside, Riverside, CA, USA.
Point of contact: {\tt\small xwang64@gmu.edu}.}
}
\begin{document}
\maketitle
\thispagestyle{empty}
\pagestyle{empty}

%%%%%%%%%%%%%%%%%%%%%%%%%%%%%%%%%%%%%%%%%%%%%%%%%%%%%%%%%%%%%%%%%%%%%%%%%%%%%%%%
\begin{abstract}
This paper studies the problem of Cooperative Localization (CL) for multi-robot systems, where a group of mobile robots jointly localize themselves by using measurements from onboard sensors and shared information from other robots. We propose a novel distributed invariant Kalman Filter (DInEKF) based on the Lie group theory, to solve the CL problem in a 3-D environment. Unlike the standard EKF which computes the Jacobians based on the linearization at the state estimate,  DInEKF 
defines the robots' motion model on matrix Lie groups and offers the advantage of state estimate-independent Jacobians. This significantly improves the consistency of the estimator. Moreover, the proposed algorithm is fully distributed, relying solely on each robot's ego-motion measurements and information received from its one-hop communication neighbors. The effectiveness of the proposed algorithm is validated in both Monte-Carlo simulations and real-world experiments. The results show that the proposed DInEKF outperforms the standard distributed EKF in terms of both accuracy and consistency.
\end{abstract}

%%%%%%%%%%%%%%%%%%%%%%%%%%%%%%%%%%%%%%%%%%%%%%%%%%%%%%%%%%%%%%%%%%%%%%%%%%%%%%%%
\section{INTRODUCTION}
For multi-robot systems, Cooperative localization (CL) is a commonly used method to precisely determine the robots' poses (position and orientation), especially in GPS-denied environments, e.g., indoor and underwater.
CL is crucial for many multi-robot applications, such as target tracking, formation control, surveillance and reconnaissance, and mapping. In CL, a group of communicating robots perceive and measure each other based on the equipped sensors, and cooperatively estimate their states by using their own measurements and the shared information. Compared with single-robot localization which is well-studied, multi-robot CL is more challenging. It requires not only the ego-motion information of each robot but also the relative robot-to-robot measurements. 

CL can be solved with both centralized and distributed methods. The centralized method integrates the motion information and sensor measurements of each robot, and performs computation in a fusion center \cite{Hnfh2023}. While centralized methods achieve optimal localization performance, they are susceptible to single-node failures and demand heavy communication and computational resources. In contrast, distributed algorithms for CL have been developed and gained more attention due to their scalability, robustness, and efficiency \cite{Pzhu2021-TCST, YanJia2024}. Among the existing distributed CL algorithms, EKF-based \cite{roumeliotis2002distributed} or its equivalent information filter-based \cite{SSS2023, pengxiang2019} algorithms are the most widely employed method due to their efficiency and accuracy. EKF-based CL algorithms generally involve two steps \cite{Calc2013}: In the first step, each robot runs a local estimator to obtain its own pose estimate using its ego-motion information and sensor measurements. In the second step, each robot communicates with other robots in the team and uses relative measurements, or so-called robot-to-robot measurements, to improve the local estimate by fusing the shared information. 
\begin{figure}[t]
    \vspace{-2ex}
    \centering
    \includegraphics[width=1.0\linewidth]{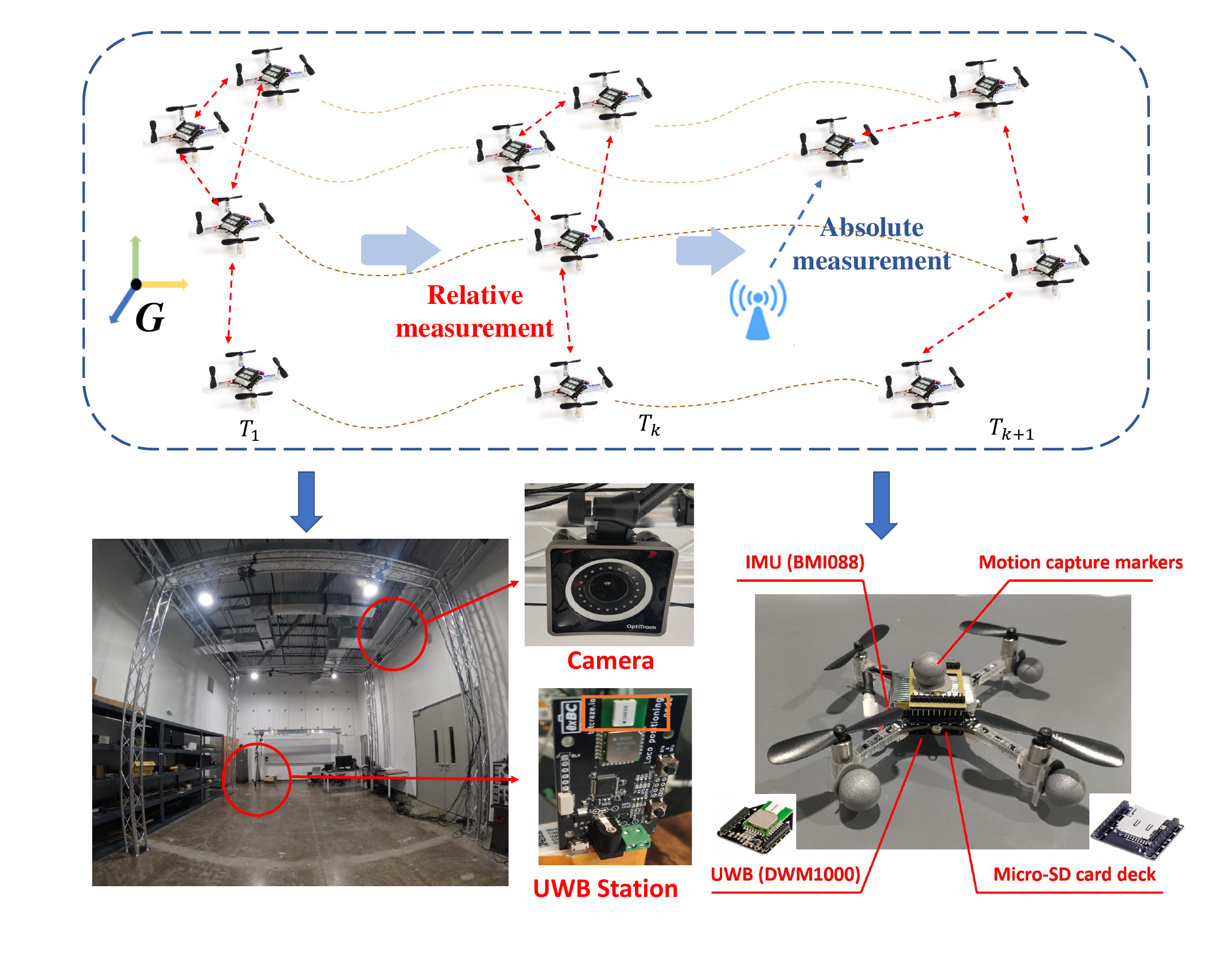}
    % \caption{Experimental testbed for multi-robot CL: four Crazyflie 2.1 nano quadrotors with expansion decks and motion capture markers attached. Each robot contains an onboard IMU, a UWB tag, and an SD-card for data collection. The groundtruth trajectory of each robot is obtained by the motion capture.}
    \caption{Cooperative Localization of a multi-robot system in an indoor environment with four Crazyflie 2.1 nano drones, based on the proposed Distributed Invariant Extended Kalman Filter (DInEKF): Each robot performs localization using only the local information (ego-motion measurements and absolute measurements), and shared information from other robots (relative measurements). The performance is evaluated using the ground truth obtained from the motion capture.}
    \label{Fig_Exp_pl}
    \vspace{-3ex}
\end{figure}
It is worth noting that in CL, the pose estimations of each robot become correlated when robots exchange information with each other and use the shared information to update the local estimates \cite{Calc2013}. In centralized or centralized-equivalent distributed EKF \cite{roumeliotis2002distributed, kia2014centralized}, the cross-correlation among robots can be memorized by a fusion center or an assistant server. However, in a fully distributed setting, the correlation is unknown and intractable \cite{Calc2013}. Naively ignoring the correlations and fusing the correlated estimates yields an \textit{inconsistent} estimator that may reduce the performance or even diverge. 
An inconsistent estimator means that the estimated covariance tends to be overconfident and lacks the uncertainty information of an estimate. To address this issue, fusion strategies such as covariance intersection (CI) \cite{Calc2013, SSS2023, pengxiang2019}, split covariance intersection (SCI) \cite{YanJia2024}, and inverse covariance intersection (ICI) \cite{NOACK2017} have been widely used to handle the unknown correlations and ensure the consistency of the fused estimates. 

Although CI-based EKF shows good performance in solving CL, the consistency of these methods cannot be theoretically guaranteed. The CI algorithm produces a consistent fused estimate only when the individual estimates are consistent. Accordingly, most CI-based methods compute the fused estimate by implicitly assuming the consistency of each local estimate, which, however, is not always the case. This is because the EKF itself has inconsistency issues as well, which has been well-studied from the perspective of observability \cite{huang2011observability}. In EKF, the Jacobian of the linearized dynamics and the measurement are functions of the estimated states, which causes the standard EKF-based CL algorithm to have an unobservable subspace of lower dimension than that of the underlying CL system. The dimension reduction causes the unobservable states to become observable and gain spurious information that undermines the consistency. Inspired by these changes, observability-constrained (OC) \cite{HJA2014}, first-estimates Jacobian (FEJ) \cite{Ghfej2008, Ghp2010}, and Kalman decomposition-based \cite{Hnfh2023} approaches have been developed to maintain the observability property of the original CL system by adding artificial constraint into the estimators. 

Instead of directly manipulating the Jacobians or linearization points, the manifold theory, especially the Matrix Lie group representation for robot state, has become increasingly popular since the estimation error is invariant under the action of Matrix Lie group \cite{Bon2009}. The theory of invariant observer \cite{Bar2017} leads to the development of the Invariant EKF (InEKF), which has been widely applied to single-robot state estimation \cite{RHM2020, Pavl2021}, navigation \cite{PER2021} and SLAM \cite{YYL2022, GEI2022, HeoS2018, Wkz2017, Bro2018}. Li et al. \cite{Liang2021} introduced a SCI-based fusion method for multi-robot localization on Lie group with unknown correlations. InEKF is extended to a distributed fashion \cite{Xuz2023} for the first time to solve the 3-D target tracking problem in a static sensor network. However, the existing distributed InEKF explicitly assumes that the sensors' positions are given as prior information, and is inapplicable for CL, since the robot team we aim to localize naturally forms a mobile robot network. Although approaches to solving the distributed CL problem have been well-established, all these algorithms operate within the vector space, which only consider the 2-D environments \cite{Pzhu2021-TCST, SSS2023} or 3-D environments with only simple rotations.

\noindent\textbf{Statement of contribution:}
In this paper, we propose a novel CL algorithm based on Distributed Invariant EKF (DInEKF). The algorithm is derived from the matrix Lie group, and the resulting error dynamics exhibit a log-linear property, which enhances the consistency of the CL algorithm. The key contributions of our approach are as follows:\\
\noindent~$\bullet$ We formulate and develop DInEKF for solving multi-robot CL problems in 3-D environments based on InEKF. The consistency of the proposed method is guaranteed by utilizing the property of the invariant error of the matrix Lie group.\\
\noindent~$\bullet$ To design a fully distributed framework, we extend the CI algorithm to fuse information associated with the Lie group in mobile robot networks for the first time. Each robot only uses its local information and shared information from the one-hop communicating neighbors for jointly estimating the state. Moreover, it supports generic robot dynamic, and measurement models.\\
\noindent~$\bullet$ We conduct extensive Monte-Carlo simulations and real-world experiments to validate the performance of the proposed method. Results show that our method can improve the performance in both accuracy and consistency as compared to the existing methods.

\section{Preliminaries and Notations}
\subsection{Notation and Definitions}
Let $\mathbf I_{r}$ denote the $r\times r$ identity matrix; $\mathbf{0}_{m\times n}$ denote the $m\times n$ zeros matrix; $\mathbf{Tr}(\cdot)$ denote the trace of a matrix. 
When applied to a set, $|\cdot|$ denotes the cardinality. We use $\mathbf q\in\mathbb{R}^r$ to represent a vector of dimension $r$ with all real entries. Given a $3\times 1$ vector $\mathbf q=\begin{bmatrix} q_1&q_2&q_3 \end{bmatrix}^\top$, its skew-symmetric matrix is defined as
\begin{align}
\lfloor\mathbf q \times\rfloor&=
\begin{bmatrix}
0&-q_3&q_2\\q_3&0&-q_1\\-q_2&q_1&0
\end{bmatrix}
\end{align}

Let $\mathbb{G}^t(\mathcal{V},\mathcal{E}^t)$ be a directed graph characterizing the time-varying communication within a multi-robot system, where $\mathcal{V}$ is the robot set and $\mathcal{E}^t$ is the edge set. If $(j, i)\in\mathcal{E}^t, j\neq i$, it means that robot $i$ can communicate with robot $j$ at time step $t$. 
%if robot $j$ is in its communication range. 
The neighbor set is defined as $\mathcal{N}_i^t=\{j | (j,i)\in\mathcal{E}^t\}$. 
% We always assume self-loop $(i,i)\in\mathcal{E}^t$ and $i\in\mathcal{N}_i^t$ for all $i\in\mathcal{V}$ and $t=1,2,\cdots$.
In addition, each robot can detect the relative positions of other robots once they are within the sensing range, which is the robot-to-robot measurement. For simplicity, we assume that for each robot, the communication range is larger than its sensing range, i.e., when robot $i$ detects robot $j$, robot $i$ can receive the information broadcast by robot $j$ as well. 

\subsection{Problem Formulation}
Consider a multi-robot system in a 3-D environment with $m$ robots whose states are unknown and need to be estimated. Let $G$ and $L_i$ denote the global frame and robot $i$'s local frame, respectively.
The state of each robot at time $t$ can be represented as a matrix
\begin{align}\label{eq_sv}
    \mathbf {x}_i^t=(^{L_{i}}_G \mathbf R^t, ^G {\mathbf v_i^t}, ^G {\mathbf p_i^t})\in\mathbb{R}^{3\times 5}
\end{align}
which contains the robot's 3-D position $^G {\mathbf p_i}\in\mathbb{R}^3$, orientation $^{L_i}_G \mathbf R\in SO(3)$ from the global frame to robot's local frame, and velocity $^G {\mathbf v_i}\in \mathbb{R}^3$.

We assume that each robot $i$ is equipped with an inertial measurement unit (IMU) sensor to measure its ego-motion information for ego-motion measurement, and exteroceptive sensors such as Ultra-Wideband (UWB) for intermittently obtaining absolute measurements for robot's own pose and relative robot-to-robot measurements with other teammates. 
The dynamic model of each robot is given as
\begin{align}\label{eq_rbdyn}
{^{L_i}_G {\dot {\mathbf R}}^\top}&={^{L_i}_G {\mathbf R}^\top}\lfloor(\bm{\omega}_i-\mathbf n_{\omega_i})\times\rfloor\nonumber\\
^G {\dot {\mathbf p_i}}&={^G {\mathbf v_i}}\nonumber\\
^G {\dot {\mathbf v_i}}&={^{L_i}_G {\mathbf R}^\top}(\mathbf a_i-\mathbf n_{a_i})+{^G \mathbf g}
\end{align}
where $\bm \omega_i, \mathbf a_i\in\mathbb R^3$ denote the robot's angular rate and acceleration with respect to its own frame $L_i$, respectively. $\mathbf n_{\omega_i}$ and $\mathbf n_{a_i}$ are the corresponding white Gaussian noises.
$^G \mathbf g\in \mathbb R^3$ is the gravity acceleration vector. 
Let $\mathbf z_{ij}^t$ denote the relative pose measurement with respect to robot $i$'s local frame at timestep $t$ between robot $i$ and robot $j$ for all $j\in\mathcal{N}_i^t$ given by
\begin{align}\label{eq_m_model}
\mathbf z_{ij}^t&=\mathbf{h}_{ij}^t(\mathbf x_i^t, \mathbf x_j^t)+\bm \eta_{ij}^t
\end{align}
%\marginA{There is confusion on how z is defined. in i or j's local frame.}
which is a function of robot $i$'s state $\mathbf x_i^t$ and robot $j$'s state $\mathbf x_j^t$. $\mathbf n_{ij}^t$ is the measurement noise with zero-mean Gaussian distribution $\bm \eta_{ij}^t\sim\mathcal{N}(\mathbf 0, \mathbf Q_{ij})$. In this work, we allow $\mathbf h_{ij}^t$ to be any measurement model related to the relative pose, such as distance-bearing measurements or distance-only measurements.
If accessible, robot $i$ receives the absolute measurement $\mathbf z_i^t$ of the state which is modeled by
\begin{align}
\mathbf z_i^t&=\mathbf h_i^t(\mathbf x_i^t)+\bm\eta_i^t   
\end{align}
with measurement noise $\bm\eta_i^t\sim\mathcal{N}(\mathbf 0, \mathbf Q_{i})$. It is worth noting that robot $i$ may not able to obtain any relative or absolute measurements at some timestep $t$. Furthermore, we assume that the measurement noises among all the sensors are mutually uncorrelated for different robots. 
The objective of this work is for each robot $i$ to compute its own state estimate by using its local measurements, and information from its one-hop communicating neighbors if accessible at the current timestep $t$.
% Each robot has communication devices that allow information exchange for each robot $i$ within its communicating neighbor set $\mathcal{N}_i^t$.

\section{Lie Group Theory}
In this section, we briefly introduce the matrix Lie group theory \cite{Bar2017} used to derive our algorithm.
\subsection{Matrix Lie Group and Lie Algebra}
A matrix Lie group $\mathcal G\in\mathbb R^{N\times N}$ is a subset of square invertible matrices with the following three properties holding:
\begin{align}
\mathbf I_N\in \mathcal G;\nonumber\\
\forall X\in \mathcal G,X^{-1}\in \mathcal G;\nonumber\\
\forall X_1,X_2\in \mathcal G,X_1 X_2\in \mathcal G
\end{align}
The corresponding Lie algebra of a matrix Lie group $\mathcal G$ denoted as $\mathfrak{g}$, is a vector space with the same dimension as $\mathcal{G}$. For any element $\bm\xi$ in $\mathfrak g$ denoted as $\bm\xi\in\mathbb R^{dim (\mathfrak{g})}$, it can be transformed to its Lie group using the exponential map $\exp_{\mathcal G}(\cdot):\mathbb{R}^{dim \mathfrak{g}}\rightarrow \mathcal G$ as
\begin{align}
\exp_{\mathcal G}(\bm\xi)=\exp(\bm\xi^\wedge)=\sum_{k=1}^{\infty}\frac{\bm\xi^\wedge}{k!}    
\end{align}
where the "hat" operator $(\cdot)^\wedge$ denotes the linear mapping $\mathbb{R}^{dim \mathfrak{g}}\to\mathfrak{g}$, that transform the element in the Lie algebra $\mathfrak g$ to the corresponding matrix form.
The inverse function of the exponential map, that is, the logarithm map $\log_{\mathcal G}:\mathcal G\rightarrow \mathbb R^{dim (\mathfrak{g})}$, can be defined as
\begin{align}
\log_{\mathcal G}(\exp_{\mathcal G}(\bm\xi))=\bm\xi    
\end{align}

Let $X^t\in \mathcal G$ be the state of a system at timestep $t$, and the dynamic model of the system on the matrix Lie Group can be represented as
\begin{align}
\frac{d}{dt}X^t=f_{u^t}(X^t)
\end{align}
where $u^t$ is the input of the system. The system $f_{u^t}$ is said to be group affine if it satisfies:
\begin{align}
f_{u^t}(X_1 X_2)=f_{u^t}(X_1)X_2+X_1 f_{u^t}(X_2)-X_1 f_{u^t}(\mathbf I_N)X_2
\end{align}
for all $t>0$ and $\mathbf I_N, X_1, X_2 \in \mathcal G$.
Further, let $\hat X^t$ denote the estimate of the state $X^t$, and the right invariant error between $X^t$ and $\hat X^t$ is given by
\begin{align}\label{eq_right_inv_er}
\bm\eta^t=X^t(\hat X^t)^{-1}
\end{align}
Note that when $X^t=\hat X^t$, the right invariant error reduces to the identity. Based on the log-linear property of the error $\bm\eta_t$ \cite[Theorem 2]{Bar2017}, if the dynamics $f_{u^t}$ is group affine, the right invariant error $\bm\eta^t$ between $X^t$ and $\hat X^t$ for all $t>0$ can be computed by
\begin{align}
\bm\eta^t=\exp_{\mathcal G}(\bm\xi^t)\quad \text{given}\quad \bm\eta^0=\exp_{\mathcal G}(\bm\xi^0)  
\end{align}

\subsection{State Representation in Matrix Lie Groups}
The state of each robot $i$ at timestep $t$ in equation \eqref{eq_sv} forms a natural Lie group representation through the group ${SE_2(3)}$ given as:
\begin{align}\label{eq_state_mr}
 X_i^t=
 \begin{bmatrix}
 (^{L_i} _G \mathbf R^t)^\top& ^G {\mathbf v_i^t} & ^G {\mathbf p_i^t} \\
 \mathbf {0}_{1\times 3} & 1& 0\\
 \mathbf {0}_{1\times 3} & 0& 1
 \end{bmatrix}
\end{align}
Recall the definition of right invariant error in equation \eqref{eq_right_inv_er}. Let $\hat{X_i^t}$ be robot $i$'s state estimate as
\begin{align}
\hat X_i^t=
\begin{bmatrix}
(^{L_i} _G {\hat{\mathbf R}}^{t})^\top& ^G {\hat{\mathbf v}_i^{t}} & ^G {\hat{\mathbf p}_i^{t}} \\
\mathbf {0}_{1\times 3} & 1& 0\\
\mathbf {0}_{1\times 3} & 0& 1
\end{bmatrix}
\end{align}
and the right invariant estimation error can be represented as 
\begin{align}
\bm\eta_i^t&= X_i^t(\hat X_i^t)^{-1}=
\begin{bmatrix}
\Lambda_1&\Lambda_2&\Lambda_3\\
\mathbf{0}_{1\times 3}&1&0\\
\mathbf{0}_{1\times 3}&0&1
\end{bmatrix}
\end{align}
with
\begin{align}
\Lambda_1 &= (^{L_i} _G \mathbf R^t)^\top (^{L_i} _G {\hat{\mathbf R}}^t)\nonumber\\
\Lambda_2 &= {^G {\mathbf v_i^t}}-\Lambda_1 {^G {\hat {\mathbf v}}_i^t}\nonumber\\
\Lambda_3 &= {^G {\mathbf p_i^t}}-\Lambda_1 {^G {\hat {\mathbf p}}_i^t}
\end{align}
The corresponding error uncertainty vector $\bm\xi_i^t$ defined in the Lie algebra of ${SE_{2}(3)}$, denoted by ${se_2(3)}$, is give by
\begin{align}
\bm\xi_i^t=\begin{bmatrix}
(\bm\xi_{R_i}^t)^\top&(\bm\xi_{v_i}^t)^\top&(\bm\xi_{p_i}^t)^\top\end{bmatrix}^\top
\end{align}
where 
\begin{align}
(\bm\xi_i^t)^\wedge&=
\begin{bmatrix}
\lfloor\bm\xi_{R_i}^t \times\rfloor&\bm\xi_{v_i}^t&\bm\xi_{p_i}^t\\
\mathbf{0}_{2\times 3}& \mathbf{0}_{2\times 1} &\mathbf{0}_{2\times 1}
\end{bmatrix}
\end{align}

From \cite{RHM2020}, equation \eqref{eq_rbdyn} is group affine without noise, and the dynamics of $\bm\xi_i^t$ can be represented as
\begin{align}
\frac{d}{dt}\bm\xi_i^t=A_i^t\bm\xi_i^t-\text{Ad}_{\hat X_i^t} W_i^t
\end{align}
where $\text{Ad}_{\hat X_t}$ denotes the adjoint matrix of $SE_2(3)$ at $\hat X_i^t$ given as
\begin{align}
\text{Ad}_{\hat X_i^t}=
\begin{bmatrix}
(^{L_i}_G \hat {\mathbf R}^t)^\top & \mathbf{0}_{3} & \mathbf{0}_{3}\\
\lfloor^G {\hat {\mathbf v}}_i^t{\times}\rfloor (^{L_i}_G \hat {\mathbf R}^t)^\top & (^{L_i}_G \hat {\mathbf R}^t)^\top & \mathbf{0}_{3}\\
\lfloor^G {\hat {\mathbf p}}_i^t{\times}\rfloor (^{L_i}_G \hat {\mathbf R}^t)^\top& \mathbf{0}_{3} & (^{L_i}_G \hat {\mathbf R}^t)^\top
\end{bmatrix}
\end{align}
and 
\begin{align}\label{eq_A}
A_i^t&=
\begin{bmatrix}
\mathbf{0}_{3} & \mathbf{0}_{3} & \mathbf{0}_{3}\\
\lfloor^G{\bm g}{\times}\rfloor & \mathbf{0}_{3} & \mathbf{0}_{3}\\
\mathbf{0}_{3} & \mathbf{I}_{3} & \mathbf{0}_{3}
\end{bmatrix}\\
W_i^t&=
\begin{bmatrix}
(\mathbf n_{\omega_i}^t)^\top&(\mathbf n_{a_i}^t)^\top&\mathbf{0}_{1\times 3}
\end{bmatrix}^\top
\end{align}
where $A_i^t$ is time-invariant and independent on the state estimate.

\section{Proposed Algorithm}
In this section, we propose a fully distributed CL algorithm based on the matrix Lie groups. 

Let $\bar X_i^t$ and $\hat X_i^t$ be, respectively, the robot $i$'s prior state estimate and posterior state estimate of its true state $X_i^t$ at timestep $t$. According to the definition of right invariant error, we define the prior and posterior estimation error as $\bar\eta_i^t$ and $\hat\eta_i^t$, respectively, given by
\begin{align}
\bar\eta_i^t&=X_i^t(\bar X_i^t)^{-1}=\exp_G(\bar{\bm \xi}_i^t)\nonumber\\
\hat\eta_i^t&=X_i^t(\hat X_i^t)^{-1}=\exp_G(\hat{\bm \xi}_i^t)
\end{align}
where $\bar{\bm \xi}_i^t$ and $\hat{\bm \xi}_i^t$ denote, respectively, prior and posterior estimation error of each robot defined in Lie algebra $se_2(3)$. The corresponding error covariance matrices associated with $\bar{\bm \xi}_i^t$ and $\hat{\bm \xi}_i^t$ are denoted as $\bar P_i^t$ and $\hat P_i^t$, respectively.

\noindent\textbf{Propagation}: 
Suppose that at timestep $t$, each robot possesses a posterior estimate $(\hat X_i^{t-1}, \hat P_i^{t-1})$ of the previous timestep. The objective of this step is to propagate the posterior estimate to obtain a prior estimate $(\bar X_i^{t}, \bar P_i^{t})$ based on the motion model. The derivation of this step is provided in the Appendix as it is standard. 

\noindent\textbf{Local measurement update}:
After obtaining the prior estimate $(\bar X_i^{t}, \bar P_i^{t})$, each robot $i$ uses its local absolute measurement $\mathbf z_i^t$, if accessible, to compute an intermediate estimate, denoted as $(\check X_{i}^t, \check P_{i}^t)$. The measurement residual after linearization at $\bar X_i^{t}$ is given by
\begin{align}
\bar{\mathbf z}_i^t=\mathbf C_i^t \bar{\bm \xi}_i^t+\bm \eta_i^t
\end{align}
where $\bar{\mathbf z}_i^t={\mathbf z}_i^t-\mathbf h_i(\bar {\mathbf x}_i^t)$ and $\bar {\mathbf x}_i^t$ is the general matrix representation (\textbf{cf.} equation \eqref{eq_state_mr}) of $\check X_i^t$. $\mathbf C_i^t$ denotes the measurement Jacobian with respect to the right invariant error. Note that we don't specify the measurement model $\mathbf h_i^t$ and allow it to be any kind of state measurement. Next, we use the information above to update the prior estimation according to the general EKF framework. The Kalman gain $\mathbf K_i$ of robot $i$ is given by
\begin{align}
\mathbf K_i&=\bar P_i^t {\mathbf C_i^t}^\top(\mathbf C_i^t \bar P_i^t {\mathbf C_i^t}^\top+\mathbf Q_i)^{-1}
\end{align}
After obtaining the Kalman gain, the state correction, denoted as $\bar{\bm \varepsilon}_{i}^t$, can be computed as
\begin{align}\label{eq_inm_cp}
\bar{\bm \varepsilon}_{i}^t&= \mathbf K_i\bar{\mathbf z}_{i}^t
\end{align}
Then $\bar{\bm \varepsilon}_{i}^t$ is used to update the prior estimate $\bar X_i^t$ for obtaining the intermediate estimate $\check X_i^t$ as
\begin{align}\label{eq_inm_s}
\check X_{i}^t&=\exp_{\mathcal G}(\bar{\bm \varepsilon}_{i}^t)\bar X_i^t
\end{align}
and the covariance $\check P_i^t$ is given as
\begin{align}\label{eq_inm_p}
\check P_i^t&=(\mathbf I_{9}-\mathbf K_i \mathbf C_i^t) \bar P_i^t   
\end{align}
% Then the intermediate state estimate $\check X_i^t$ is obtained by using $\bar{\bm \varepsilon}_{i}^t$ to update $\bar X_i^t$ as
% \begin{align}\label{eq_inm_s}
% \check X_{i}^t&=\exp_G(\bar{\bm \varepsilon}_{i}^t)\bar X_i^t
% \end{align}
% and the corresponding covariance estimate $\check P_i^t$ is given by
% \begin{align}\label{eq_inm_p}
% \check P_{i}^t&=\begin{bmatrix}
% (\bar P_{i}^t)^{-1} +\mathbf S_{i}^t
% \end{bmatrix}^{-1}
% \end{align}
% with 
% \begin{align}\label{eq_inm_cp}
% \mathbf S_{i}^t&=(\mathbf C_i^t)^\top (\mathbf{Q}_{i})^{-1}\mathbf C_i^t
% \end{align}
Now define $\check {\bm \xi}_i^t$ as the new error vector in $se_2(3)$ of robot $i$ which satisfies
\begin{align}
\exp_{\mathcal G}(\check {\bm \xi}_i^t)&=X_i^t (\check X_i^t)^{-1}
\end{align}
Note that if no absolute measurements is received at current time, the intermediate estimate $(\check X_i^t, \check P_i^t)$ will be exactly the prior estimate $(\bar X_i^t, \bar P_i^t)$.

\noindent\textbf{Relative measurement update}: If robot $i$ has access to the relative robot-to-robot measurement $\mathbf z_{ij}^t$ at the current time $t$,
the robot will use this measurement to compute a posterior estimate denoted as $(\hat X_i^{t}, \hat P_i^{t})$, for improving the accuracy of $(\check X_i^{t}, \check P_i^{t})$.

Specifically, when robot $i$ detects robot $j$ for $j\in\mathcal{N}_i^t$, robot $i$ obtains the robot-to-robot measurements $\mathbf z_{ij}^t$ and receives information broadcast by robot $j$, which contains the robot $j$'s current intermediate estimate $(\check X_j^{t}, \check P_j^{t})$. 
Linearize the measurement $\mathbf z_{ij}^t$ at $\check X_i^t$ and $\check X_j^t$ to compute the measurement residual
\begin{align}\label{eq_z_res}
\bar{\mathbf z}_{ij}^t&=\mathbf{H}_{i}^t \check{\bm \xi}_{i}^t+ {\mathbf H}_{j}^t \check{\bm \xi}_{j}^t+\bm\eta_{ij}^t
\end{align}
where $\bar{\mathbf z}_{ij}^t=\mathbf z_{ij}^t-\mathbf h_{ij}^t(\check {\mathbf x}_i^t, \check {\mathbf x}_j^t)$. Similarly, $\check {\mathbf x}_i^t$ and $\check {\mathbf x}_j^t$ be, respectively, robot $i$'s and robot $j$'s intermediate estimate in its general matrix form. $\mathbf{H}_{i}^t$ and $\mathbf{H}_{j}^t$ represents the measurement Jacobian of $X_i^t$ and $X_j^t$, respectively.
By defining $\mathbf e_{ij}^t={\mathbf H}_{j}^t \check{\bm \xi}_{j}^t+{\bm\eta}_{ij}^t$, equation \eqref{eq_z_res} can be rewritten as
\begin{align}\label{eq_z_res_spl}
\bar{\mathbf z}_{ij}^t&=\mathbf{H}_{i}^t \check{\bm \xi}_{i}^t+ \mathbf e_{ij}^t 
\end{align}
and the covariance of $\mathbf e_{ij}^t$, denoted as $\mathbf R_{ij}^t$, is given by 
\begin{align}\label{eq_zl_cov}
\mathbf R_{ij}^t=\mathbf{Q}_{ij}+{\mathbf H}_{j}^t\bar P_j^t({\mathbf H}_{j}^t)^\top
\end{align}
which includes the uncertainty of robot $j$'s position estimate. Next define the relative correction pair $\mathbf S_{ij}^t$ and $\mathbf y_{ij}^t$ as
\begin{align}\label{eq_inm_cp}
\mathbf S_{ij}^t&=(\mathbf H_i^t)^\top (\mathbf{R}_{ij}^t)^{-1}\mathbf H_i^t\nonumber\\
\mathbf y_{ij}^t&=
(\mathbf H_i^t)^\top (\mathbf{R}_{ij}^t)^{-1}(\bar{\mathbf z}_{ij}^t+\mathbf{H}_{i}^t \check {\mathbf x}_i^t)
\end{align}
Then the task is to compute the posterior estimate $(\hat X_i^t, \hat P_i^t)$ from the available correction pair $\mathbf S_{ij}^t$, $\mathbf y_{ij}^t$ for $j\in\mathcal{N}_i^t$ with the intermediate estimate $(\check X_i^t, \check P_i^t)$. A naive way for computing $(\hat X_i^t, \hat P_i^t)$ is to directly fuse $(\check X_i^t, \check P_i^t)$ with all the $\mathbf S_{ij}^t$, $\mathbf y_{ij}^t$ based on EKF as
\begin{align}\label{eq_inm}
\hat P_{i}^t&=\begin{bmatrix}
(\check P_{i}^t)^{-1} + \sum_{j\in\mathcal{N}_i^t}\mathbf S_{ij}^t
\end{bmatrix}^{-1}\nonumber\\
\hat X_{i}^t&=\exp_{\mathcal G}(\check{\bm \varepsilon}_{i}^t)\check X_i^t
\end{align}
where $\check{\bm \varepsilon}_{i}^t$ denotes the state correction in $se_2(3)$ given as
\begin{align}
\check{\bm \varepsilon}_{i}^t=\hat P_{i}^t \sum_{j\in\mathcal{N}_i^t}\mathbf y_{ij}^t   
\end{align}
Note that equation \eqref{eq_inm} implicitly assumes that the estimates $\check X_i^t$ and $\check X_j^t, j\in\mathcal{N}_i^t$ are mutually uncorrelated. However, in practice, this is not the case. Although the relative measurement noises are mutually uncorrelated, the state estimate between robot $i$ and robot $j$ is correlated. For example, when robot $i$ exchanges information with robot $j$ and uses robot $j$'s estimate to update its own estimate at timestep $t$, their estimates become correlated from $t+1$ and the cross-correlation will be propagated in the following time. The correlation is generally unknown and intractable in distributed settings. Naively fusing the correlated information makes the estimate overconfident and causes inconsistency in the estimator. 

To generate a consistent posterior estimate, we apply the CI-EKF algorithm \cite{Cvio2021,Yzhang2023} to fuse the intermediate estimate and the relative correction pairs. In particular, the posterior covariance estimate $\hat P_i^t$ can be computed as
\begin{align}\label{CI_EKF_p}
\hat P_{i}^t&=\begin{bmatrix}
\alpha_{i}^t(\check P_{i}^t)^{-1} + \sum_{j\in\mathcal{N}_i^t}\alpha_{ij}^t\mathbf S_{ij}^t
\end{bmatrix}^{-1}
\end{align}
and the corresponding state estimate $\hat X_i^t$ is given as
\begin{align}
\hat X_{i}^t&=\exp_G(\check{\bm \varepsilon}_{i}^t)\check X_i^t
\end{align}
where $\check{\bm \varepsilon}_{i}^t$ is the fused state correction given as
\begin{align}\label{CI_EKF_c}
\check{\bm \varepsilon}_{i}^t=\frac{1}{\alpha_i^t}{\hat P_{i}^t}\sum_{j\in\mathcal{N}_i^t}\alpha_{ij}^t\mathbf{y}_{ij}^t
\end{align}
$\alpha_i^t$ and $\alpha_{ij}^t$ are weights of the CI, which satisfies $\alpha_i^t+\sum_{j\in\mathcal{N}_i^t}\alpha_{ij}^t=1$.
The detail derivation of equation \eqref{CI_EKF_p} and \eqref{CI_EKF_c} can be found in Appendix.

\begin{algorithm}%\label{alg}
\caption{DIKF-CL}\label{alg}
\begin{algorithmic}
\State\textbf{Propagation:} Each robot $i$ computes its prior estimation pair $({\bar X}_i^k,{\bar P}_i^k)$.
\State\textbf{Local measurement update:} Robot $i$ uses the absolute measurement $\mathbf z_i^t$ if it is available to update the prior estimate and obtain the intermediate estimate $({\check X}_i^k,{\check P}_i^k)$ using equation \eqref{eq_inm_s} and \eqref{eq_inm_p}.
\State\textbf{Relative measurement update:} If the relative measurement $\mathbf z_{ij}^t$ is accessible, robot $i$ will communicate with its neighbor $j,j\in\mathcal{N}_i^t$ and update the intermediate estimate $(\check X_i^t, \check P_i^t)$ to compute an improved posterior estimate $({\hat X}_i^t, {\hat P}_i^t)$.
\end{algorithmic}
\end{algorithm}

\begin{remark}[Consistency analysis of the proposed algorithm]
In the proposed DInEKF algorithm, each robot has a local estimator to determine its own state, and then fuses the shared information, which is mutually correlated, from other robots to correct its local estimate. Therefore, to ensure the consistency of the entire algorithm, it is imperative to guarantee the consistency at both the local estimator level and the information fusion process. The consistency of the proposed DInEKF naturally preserves the observability properties of the underlying CL system and ensures the consistency of the local estimator accordingly, since the linear error dynamics matrix $A_i^t$ in equation \eqref{eq_A} is time-invariant. Then we use the CI rule to fuse the correlated information to finally obtain the posterior estimate which is also consistent.       
\end{remark}

\section{Simulation Results}
\begin{figure*}[t]
    %\vspace{-1ex}
    \centering
    \begin{subfigure}[h]{0.32\textwidth}
        \centering
        \includegraphics[width=\textwidth]{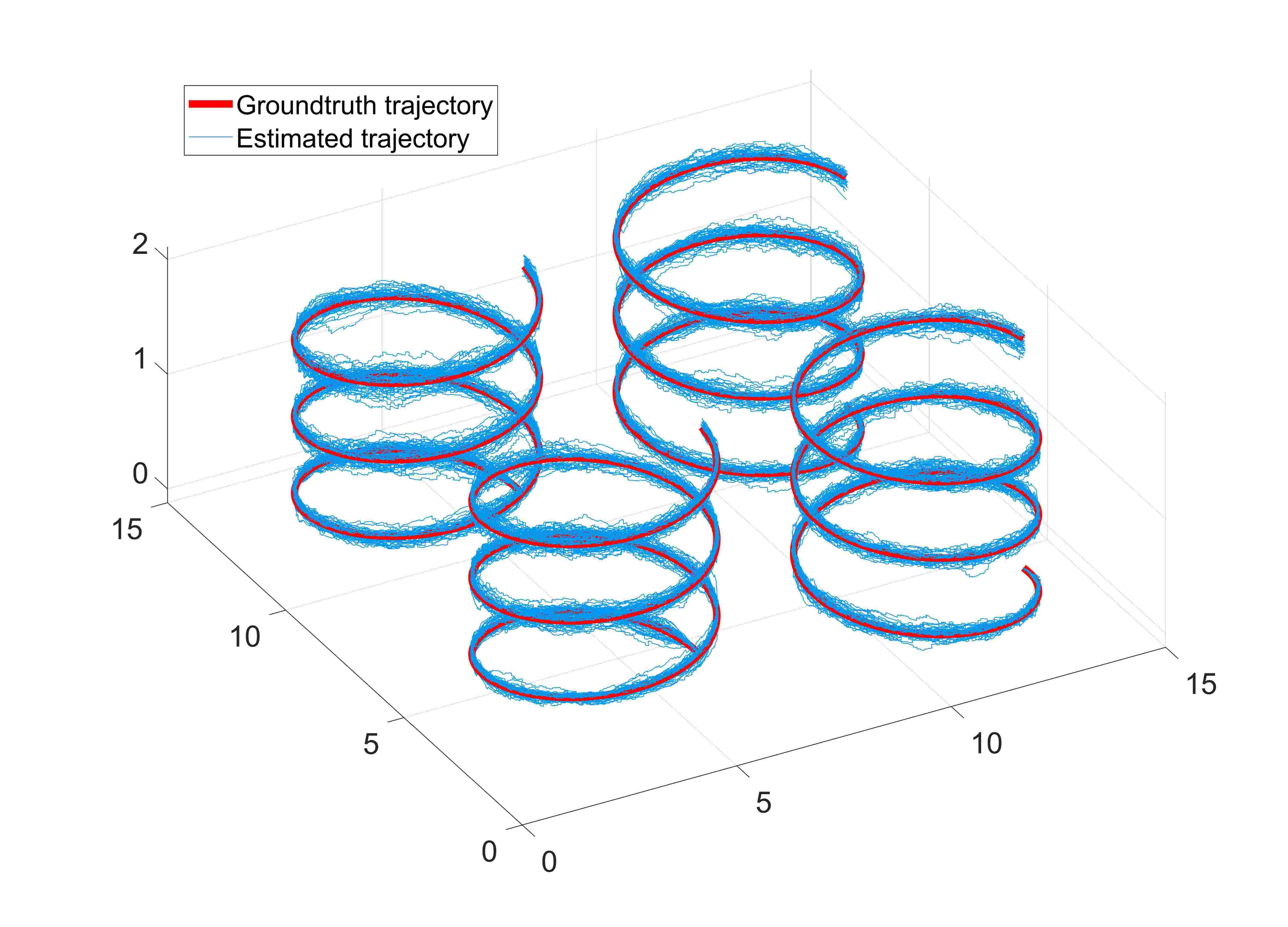}
        \caption{Trajectories 1.}
        \label{Fig_cf}
    \end{subfigure}
    \begin{subfigure}[h]{0.32\textwidth}
        \centering
        \includegraphics[width=\textwidth]{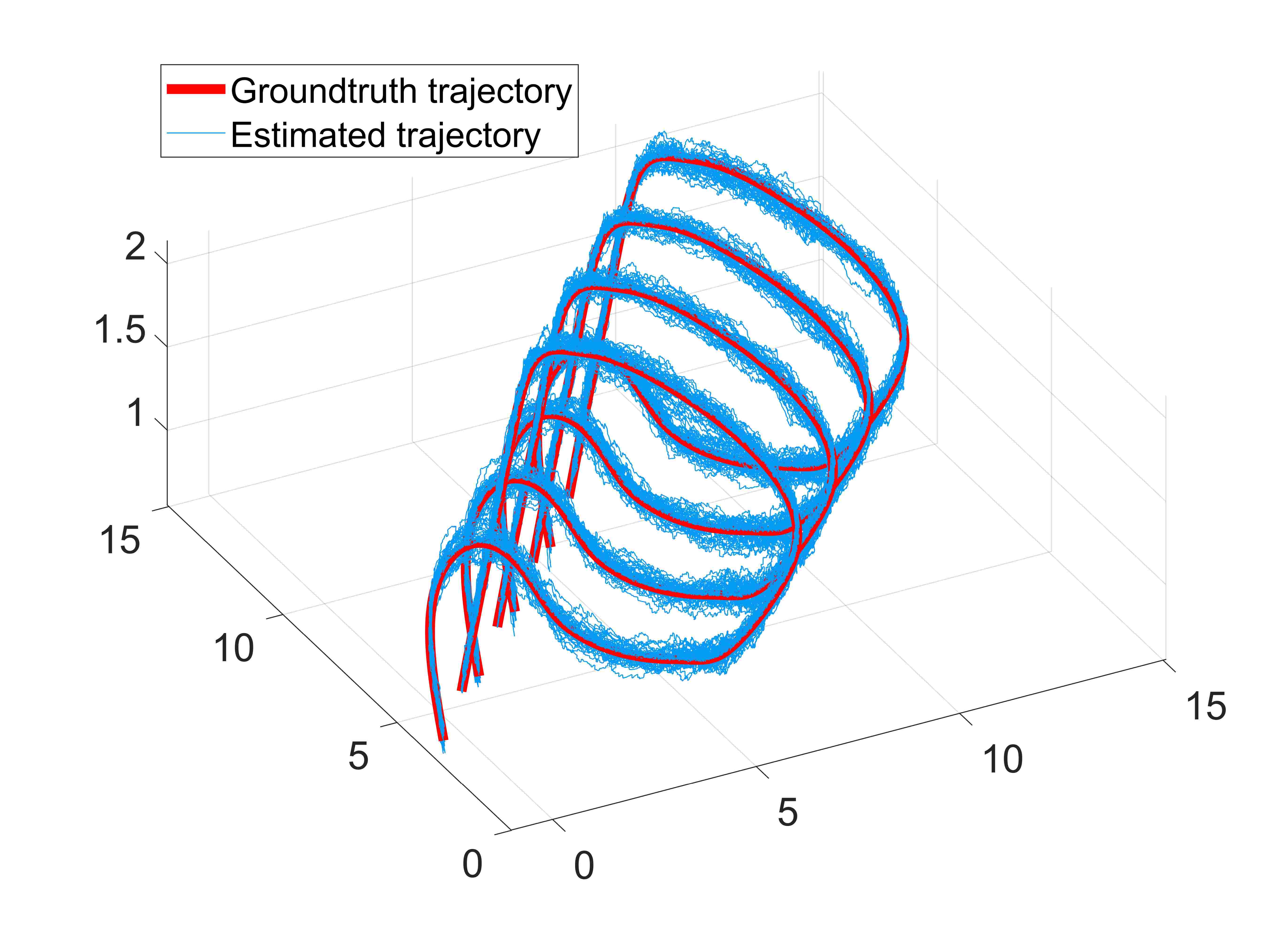}
        \caption{Trajectories 2}
        \label{Fig_mcp}
    \end{subfigure}
        \begin{subfigure}[h]{0.32\textwidth}
        \centering
        \includegraphics[width=\textwidth]{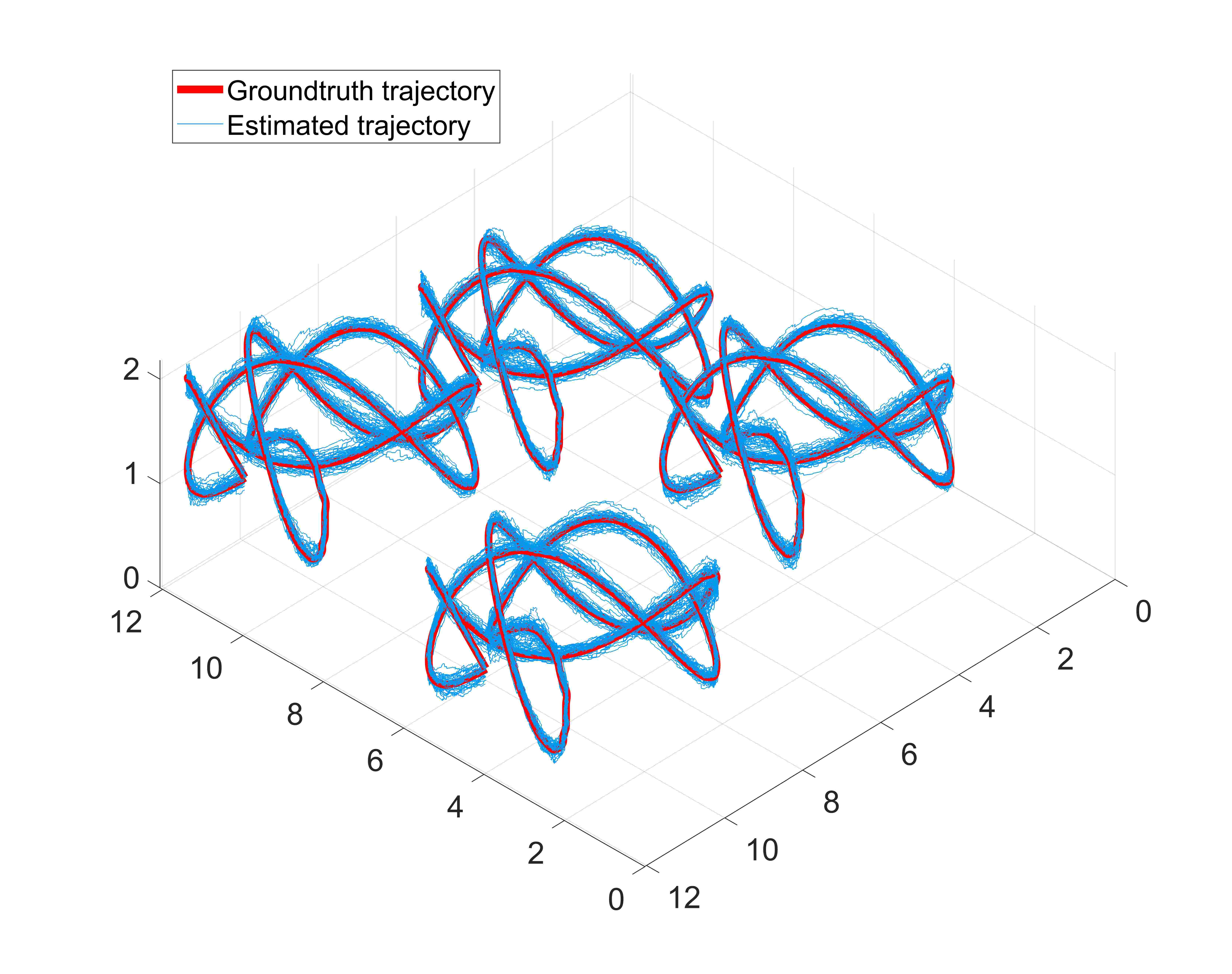}
        \caption{Trajectories 3}
        \label{Fig_mcp}
    \end{subfigure}
        \caption{CL for four drones in 3-D environments tested with three different trajectories over 50 Monte-Carlo simulations. We plot the estimated trajectories of the first 30 trails, which shows that the estimated trajectories are close to the ground truth.}
        \label{Fig_sim_pl}
\end{figure*}

In this section, we consider three environments and conduct 50 Monte Carlo simulations for each environment to validate the performance of the proposed DInEKF CL algorithm. In particular, we simulate a team of four robots in environments as depicted in Fig \ref{Fig_sim_pl}, and each robot follows a pre-designed trajectory. We assume that each robot is equipped with an IMU to measure its ego-motion, and a UWB which can obtain the relative distances to other robots and absolute distance to UWB-stations within its sensing range. The parameters used in the simulations are summarized in Table \ref{Sim_para}.
\begin{table}[h]
\vspace{-1ex}
\centering
\caption{Simulation parameters. All the algorithms and robots use the exactly same parameters as the table.}
\begin{tabular}{cccc}
   \toprule[1.2pt]
   \textbf{Parameter} & \textbf{Value} & \textbf{Parameter} & \textbf{Value}  \\
   \midrule[1.2pt]
   IMU Freq(Hz)&100& UWB Freq(HZ)&10\\
   UWB Range(m)&10 & UWB Noise(m)&0.05\\
   Gyro Noise& 2.0e-2& Gyro Bias& 3.0e-4\\
   Accel Noise& 3.0e-3& Accel Bias& 3.0e-4\\
   \bottomrule[1.2pt]
\end{tabular}
\label{Sim_para}
\end{table}

\begin{figure*}[h]
	%\vspace{-1ex}
	\centering
	\begin{subfigure}[h]{0.32\textwidth}
		\centering
		\includegraphics[width=\textwidth]{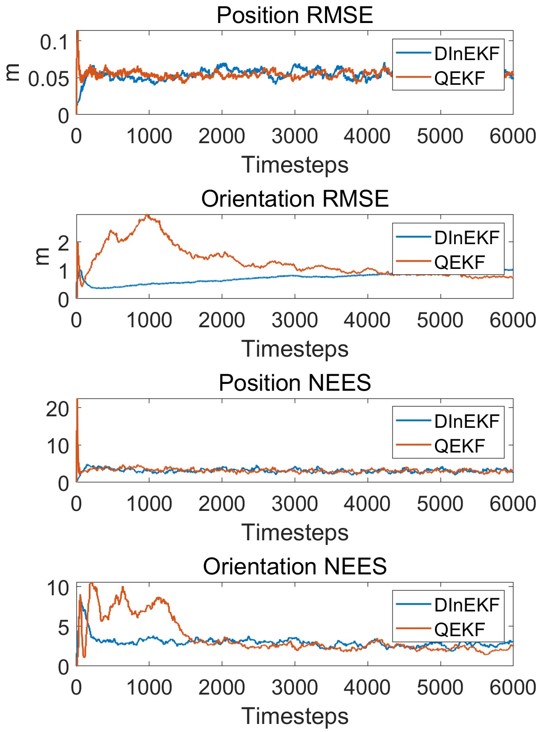}
		\caption{Trajectory set 1.}
		\label{Fig_cf}
	\end{subfigure}
	\begin{subfigure}[h]{0.32\textwidth}
		\centering
		\includegraphics[width=\textwidth]{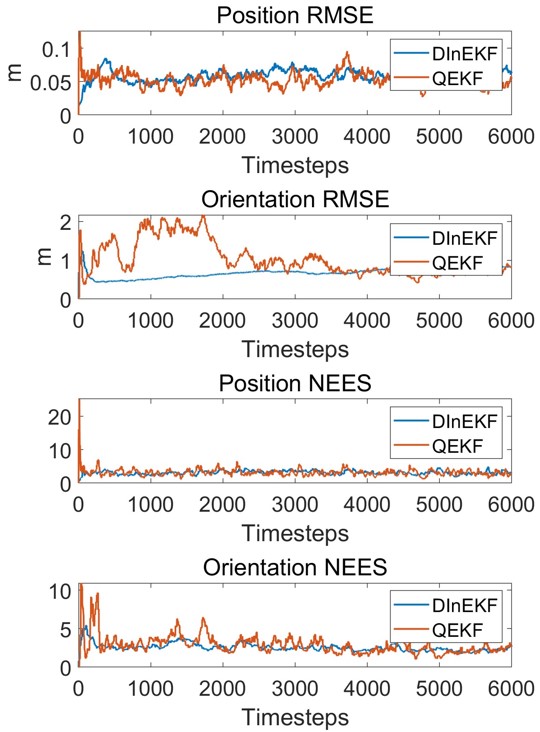}
		\caption{Trajectory set 2.}
		\label{Fig_mcp}
	\end{subfigure}
        \begin{subfigure}[h]{0.32\textwidth}
		\centering
		\includegraphics[width=\textwidth]{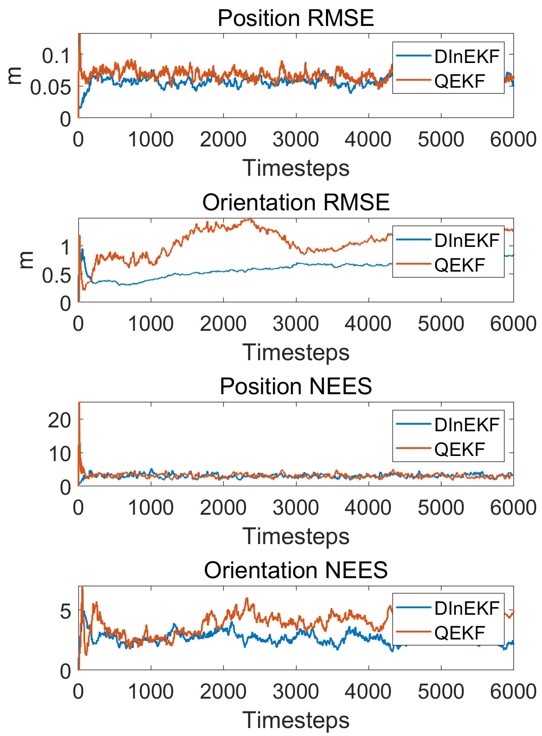}
		\caption{Trajectory set 3.}
		\label{Fig_mcp}
	\end{subfigure}
        \caption{Averaged RMSE (per robot) and NEES (per robot) results in the simulated datasets.}
        \label{Fig_err}
\end{figure*}

To further show the performance of the proposed method, we compare it with the quaternion-based distributed EKF (QDEKF) which uses decoupled error states \cite{hartley2020contact}. The Root Mean Square Error (RMSE) and Normalized Estimation Error Squared (NEES) results of each robot averaged over the Monte-Carlo simulations are given in Fig \ref{Fig_err} and Table \ref{Tab_sim}. Results show that the proposed DInEKF outperforms QDEKF in both accuracy and consistency. Specifically, the accuracy of the proposed algorithm is much higher than that of QDEKF. It should be noted that, for a consistent estimator, the NEES value ought to be close to the estimation's degrees of freedom, which is 3 in our case for both position NEES (PNEES) and orientation NEES (ONEES). The NEES of the proposed DInEKF is close to 3 in all three scenarios, whereas the QDEKF algorithm has a larger NEES value in ONEES. A larger NEES indicates that the estimator is over-confident and may yield inaccurate results, which can be observed from the orientation estimation results of QDEKF. This is because QDEKF uses the current estimate to compute the linearized Jacobians, which gains 'spurious information' along the unobservable space and leads to inconsistency. This issue has been completely eliminated by the proposed DInEKF by using the nice properties of the invariant error.

\begin{table}[t]
\vspace{-1ex}
\centering
\caption{Averaged RMSE in meter/degree of each robot over 50 Monte-Carlo simulations.}
\begin{tabular}{*{6}{c}}
  \toprule[1.2pt]
  & Estimators& PRMSE & ORMSE& PNEES & ONEES\\
  \midrule[1.2pt]
  \multirow{2}*{Traj. 1}
  &\textbf{DInEKF} &  0.057&  0.669& 3.156& 2.983\\
  &QDEKF &  0.059 &  1.097& 3.302& 3.815\\ 
  \midrule
  \multirow{2}*{Traj. 2}
  &\textbf{DInEKF} &  0.051&  0.794& 3.171& 2.936\\
  &QDEKF &  0.054&  1.169& 3.330 & 3.272\\ 
  \midrule
  \multirow{2}*{Traj. 3}
  &\textbf{DInEKF} & 0.059 & 0.662 & 3.035& 3.025\\
  &QDEKF &  0.067&  1.269 & 3.063 & 3.787\\ 
  \bottomrule[1.2pt]
  \end{tabular}
\label{Tab_sim}
\end{table}
\section{Real-world Experiments}
To further show the performance of the proposed distributed CL algorithm, we test it with four Crazyflie nano quadrotors in an indoor scenario as shown in Fig \ref{Fig_Exp_pl}. To evaluate the localization results, we use the Optitrack motion capture system in our lab to track the poses of those quadrotors as the ground truth, which contains sixteen cameras within a $10m\times 6m \times 5m$ indoor space. Each robot is equipped with an onboard IMU sensor (BMI088) to measure the ego-motion information, which is the linear acceleration and angular rate with respect to the robot's own frame. Each robot is also equipped with a UWB tag to measure the absolute distance to four fixed UWB stations. A micro-sd card deck is attached to each robot for data collection purposes, which contains a micro-sd card to record the IMU, UWB data, and the ground truth trajectories measured by the motion capture with a frequency up to 200 Hz. 

\begin{figure*}[t]
	\vspace{-1ex}
	\centering
	\begin{subfigure}[h]{0.49\textwidth}
		\centering
		\includegraphics[width=\textwidth]{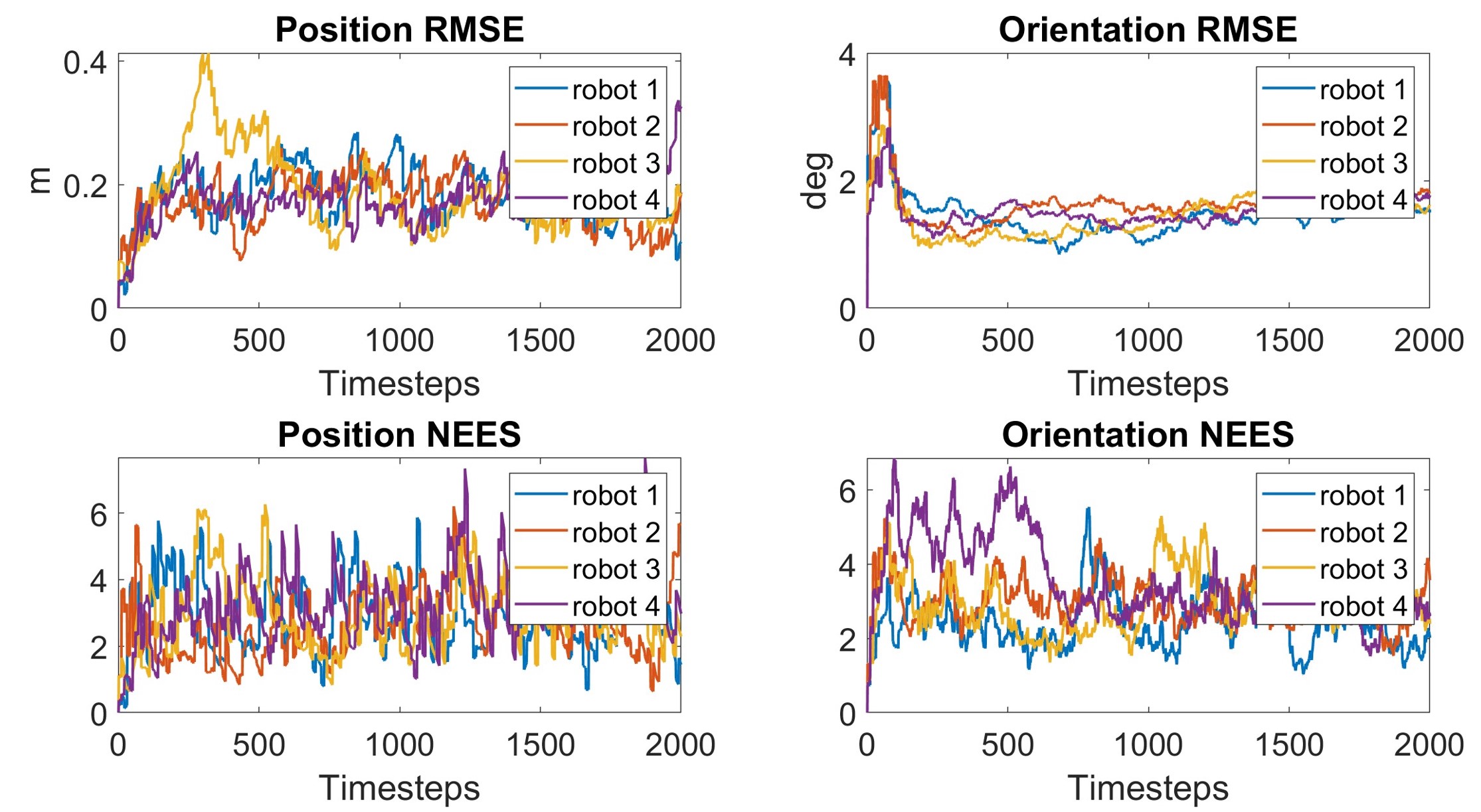}
		\caption{DInEKF}
		\label{Fig_cf}
	\end{subfigure}
	\begin{subfigure}[h]{0.49\textwidth}
		\centering
		\includegraphics[width=\textwidth]{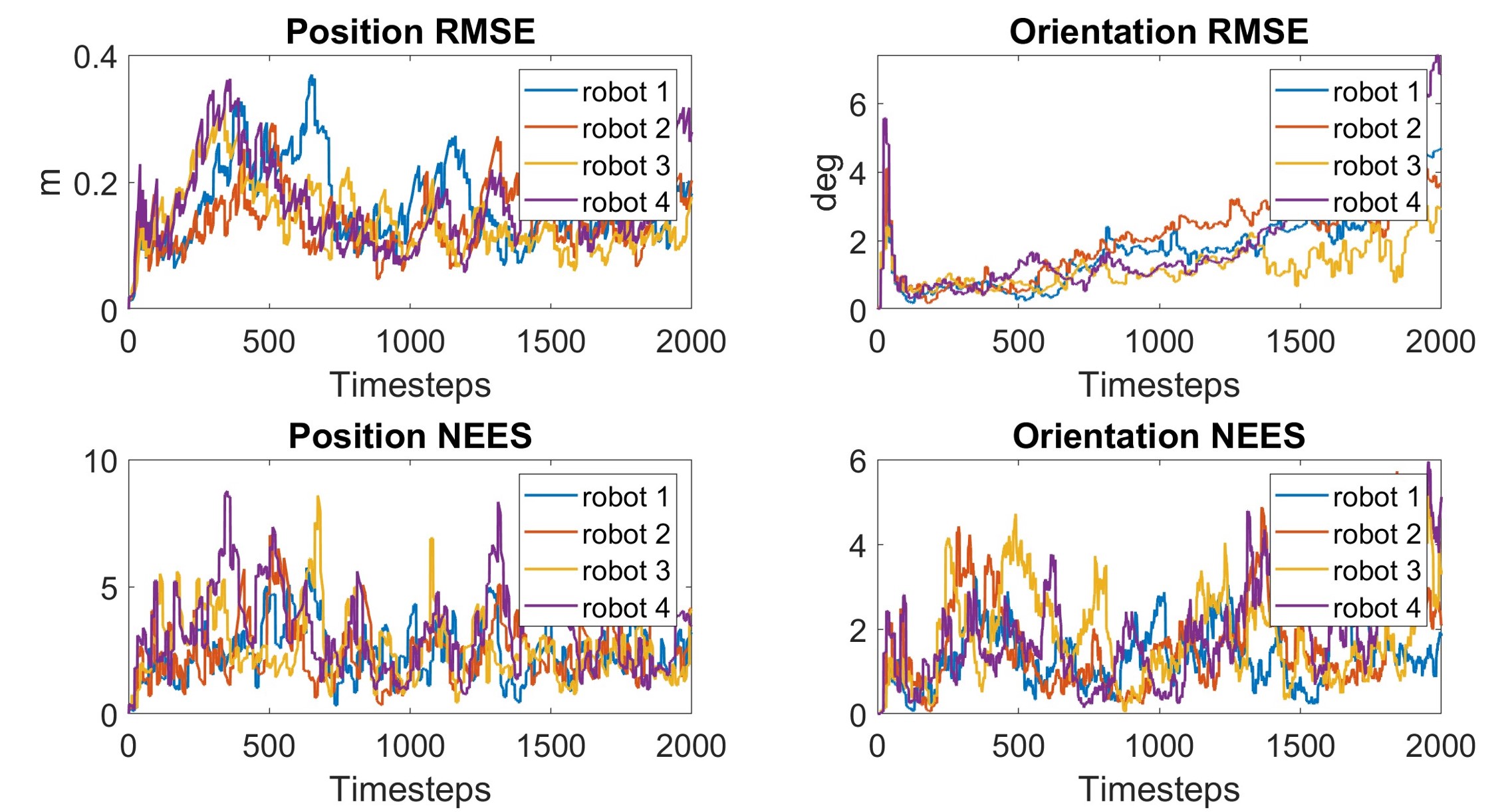}
		\caption{QDEKF}
		\label{Fig_mcp}
	\end{subfigure}
        \caption{Averaged RMSE and NEES results of dataset 1 for each robot.}
        \label{Fig_exp_r}
\end{figure*}

During the experiments, each robot moves following a randomly designed trajectory. The ego-motion information of each robot is measured by its onboard IMU at 100 Hz and recorded by the micro-sd card, and the absolute measurements are obtained by UWB. We generate the noisy relative observations with 10 Hz by computing the distances among robots based on the pose ground truth from the motion capture system.
We use the robot team to collect two experimental datasets and compare the performance of the proposed DInEKF with QDEKF as shown in Tab. \ref{Tab_exp}. It can be observed that the proposed DInEKF achieves good accuracy and consistency compared with QDEKF in both two experimental datasets. We plot the results of dataset 1 as a representative example to show the performance of individual robots as shown in Fig \ref{Fig_exp_r}.

\begin{table}[t]
\vspace{-1ex}
\centering
\caption{Averaged RMSE in meter/degree of each robot in two experimental datasets.}
\begin{tabular}{*{6}{c}}
  \toprule[1.2pt]
  & Estimators& PRMSE & ORMSE& PNEES & ONEES\\
  \midrule[1.2pt]
  \multirow{2}*{Traj. 1}
  &DInEKF &  0.22&  \textbf{2.01}& \textbf{3.27} & \textbf{3.29}\\
  &QDEKF &  \textbf{0.20}&  2.44& 3.49 & 3.41\\ 
  \midrule
  \multirow{2}*{Traj. 2}
  &\textbf{DInEKF} &0.27&  2.62&  3.34& 3.21\\
  &QDEKF &  0.35&  3.23& 3.41 & 3.66\\ 
  \bottomrule[1.2pt]
  \end{tabular}
\label{Tab_exp}
\end{table}

\section{Conclusion}
In this paper, we propose a distributed invariant EKF (DInEKF) algorithm for multi-robot CL. Based on the invariant observer theory, our DInEKF generated by the matrix Lie group automatically preserves the correct observability properties of the CL which improves the consistency. The effectiveness of our algorithm has been validated in both simulations and experiments. In the future, we will extend the proposed algorithm into SLAM problems.

\section*{Appendix}
\subsection{Discretized IMU Propagation In Lie Group}\label{A}
According to \cite{hartley2020contact}, the prior estimate $(\bar X_i^t, \bar P_i^t)$ can be propagated from $(\hat X_i^t, \hat P_i^t)$ as
\begin{align}
(^{L_i} _G {\bar{\mathbf R}}^{t})^\top  &= (^{L_i}_G {\hat{\mathbf R}}^{t-1})^\top \Gamma_0(\bm\omega_i^{t-1} \Delta t)\nonumber\\
^G {\bar{\mathbf v}_i^{t}} &= ^G {\hat{\mathbf v}_i^{t-1}} + +{^G \mathbf g}\Delta t\nonumber\\
&+(^{L_i}_G {\hat{\mathbf R}}^{t-1})^\top \Gamma_1(\bm\omega_i^{t-1} \Delta t){\mathbf a_i}^{t-1}\Delta t\nonumber\\
^G {\bar{\mathbf p}_i^{t}} &= ^G {\hat{\mathbf p}_i^{t-1}} + ^G {\hat{\mathbf v}_i^{t-1}}\Delta t +\frac{1}{2}{^G \mathbf g}\Delta t\nonumber\\
&+ (^{L}_G {\hat{\mathbf R}}_i^{t-1})^\top \Gamma_2(\bm\omega_i^{t-1} \Delta t){\mathbf a}^{t-1}\Delta t^2
\end{align}
where $\Gamma_m(\phi)$ is defined as \cite{bloesch2013state}
\begin{equation*}
\begin{aligned}
        \Gamma_m(\phi) = \left( \sum_{n=0}^\infty \frac{1}{(n+m)!}{\lfloor\phi{\times}\rfloor}^n \right),
\end{aligned}
\end{equation*} 
and $\Delta t$ denotes the time interval between timesteps $t$ and $t-1$.
The corresponding state transition matrix $\Phi(t,t-1)$, associated with $A_t$ defined in \eqref{eq_A}, can be computed as
\begin{align}
{\Phi_i(t,t-1)} = \begin{bmatrix}
\mathbf{I}_3 & \mathbf{0}_3 & \mathbf{0}_3\\
\lfloor{^G \mathbf g{\times}}\rfloor\Delta t & \mathbf{I}_3 & \mathbf{0}_3 \\
\mathbf{0}_3 & \mathbf{I}_3 \Delta t & \mathbf{I}_3 \\
\end{bmatrix}.
\end{align}
and the covariance $\bar{P}_i^t$ can be propagated from $\hat{P}_i^{t-1}$ as
\begin{align}    
    \bar{P}_i^t &= {\Phi_i(t,t-1)}\hat{P}_i^{t-1}{\Phi_i(t,t-1)} + \nonumber\\
    &{\Phi_i(t,t-1)}\mathbf O_i^{t-1}{\Phi_i(t,t-1)}\Delta t
\end{align}
where
\begin{align}    
    \mathbf O_i^{t-1} &= \text{Ad}_{{\hat{X}}_i^{t-1}} \text{cov}(W_i^{t-1})(\text{Ad}_{{\hat{X}}_i^{t-1}})^\top
\end{align}
where $\text{cov}(\cdot)$ denotes the covariance.

\subsection{CI-EKF}
Equation \eqref{CI_EKF_p} and \eqref{CI_EKF_c} can be directly derived using the CI-EKF \cite{Cvio2021,Yzhang2023}. Specifically, each robot possesses a local estimation pair $(\check X_i^t, \check P_i^t)$, relative measurement $\mathbf z_{ij}^t$, and information from its neighbor $j\in\mathcal{N}_i^t$ (\textbf{cf.}, equation \eqref{eq_z_res_spl} and \eqref{eq_zl_cov}, to update the local estimation pair. Based on our previous analysis, the shared information is not independent and mutually correlated with the local estimate as well. Directly using the update step of the standard EKF to compute the posterior estimate will yield an inconsistent result. To fuse the estimates with unknown correlation, CI-EKF update the local estimate of each robot by adopting the CI algorithm in the update step of the EKF. Given the measurement residual \eqref{eq_z_res_spl}, CI-EKF compute the covariance of the residual $\mathbf S_i^t$ as
\begin{align}\label{CI_EKF_s}
{\mathbf S_i^t}&=\sum_{j\in\mathcal{N}_i^t}\frac{1}{\alpha_{ij}^t}{\mathbf H_j^t} {\check P_i^t}^{-1}{\mathbf H_{j}^t}^\top+\mathbf{Q}_{ij}
\end{align}
the posterior covariance $\hat P_i^t$ as
\begin{align}\label{CI_EKF_pp}
\hat P_i^t=\frac{1}{\alpha_i^t}\check P_i^t-\frac{1}{({\alpha_i^t})^2}\check P_i^t {\mathbf H_i^t}^\top{\mathbf S_i^t}^{-1}{\mathbf H_i^t}\check P_i^t
\end{align}
and the state correction as
\begin{align}\label{CI_EKF_cc}
\check {\bm\varepsilon}_i^t&=\frac{1}{\alpha_i^t}\check P_i^t {\mathbf H_i^t}^\top {\mathbf S_i^t}^{-1}(\bar{\mathbf z}_{ij}^t+\mathbf H_i^t\check{\mathbf x}_i^t)
\end{align}
Then we employ the \textit{matrix inverse lemma} into equation \eqref{CI_EKF_s}-\eqref{CI_EKF_cc}, to obtain equation \eqref{CI_EKF_p} and \eqref{CI_EKF_c} as
\begin{align}
\hat P_{i}^t&=\begin{bmatrix}
\alpha_{i}^t(\check P_{i}^t)^{-1} + \sum_{j\in\mathcal{N}_i^t}\alpha_{ij}^t\mathbf S_{ij}^t
\end{bmatrix}^{-1}\nonumber\\
\check{\bm \varepsilon}_{i}^t&=\frac{1}{\alpha_i^t}{\hat P_{i}^t}\sum_{j\in\mathcal{N}_i^t}\alpha_{ij}^t\mathbf{y}_{ij}^t\nonumber
\end{align}
Equation \eqref{CI_EKF_p} and \eqref{CI_EKF_c} is indeed the information
form of the EKF, or the so-called information filter (IF), which is a dual
representation of the standard EKF. Instead of using the standard update step as equation \eqref{CI_EKF_cc} and \eqref{CI_EKF_pp}, we apply \eqref{CI_EKF_p} and \eqref{CI_EKF_c} in our algorithm due to the robustness and efficiency of IF in distributed systems \cite{JCC2017}.

\bibliographystyle{IEEEtran}
\bibliography{references}

\addtolength{\textheight}{-12cm}   % This command serves to balance the column lengths
\end{document}